\pdfoutput=1

\documentclass[11pt]{article}

\usepackage{acl}

\definecolor{oorange}{HTML}{d95f02}
\definecolor{bblue}{HTML}{7570b3}
\definecolor{ggreen}{HTML}{1b9e77}
\definecolor{ppurple}{HTML}{e37fbb}
\definecolor{lgreen}{HTML}{9CD24A}
\definecolor{yyellow}{HTML}{FFD52D}
\definecolor{ggold}{HTML}{E1BC89}
\definecolor{ggray}{HTML}{AAAAAA}

\newcommand\napafont[1]{{\usefont{T1}{damion}{m}{n}#1}}

\usepackage{times}
\usepackage{latexsym}
\usepackage[T1]{fontenc}
\usepackage[utf8]{inputenc}
\usepackage{microtype}
\usepackage{booktabs}
\usepackage{multirow}
\usepackage{amsmath}
\usepackage{graphicx}
\usepackage{pifont}
\usepackage{subcaption}

\DeclareMathOperator*{\argmax}{arg\,max}
\DeclareRobustCommand{\wine}{%
  \begingroup
  \raisebox{-0.2em}{%
  \includegraphics[height=1em]{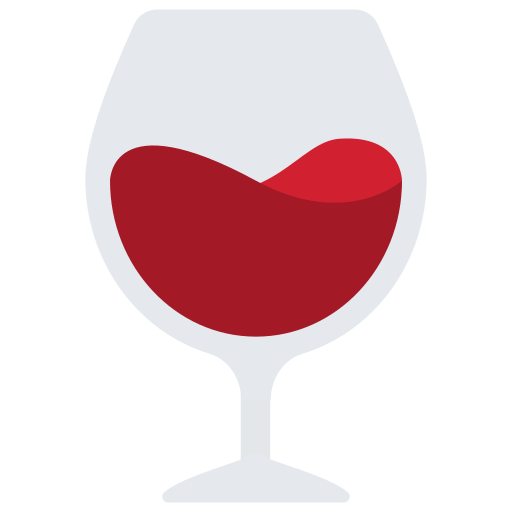}%
  }%
  \endgroup
}

\title{Noisy Pairing and Partial Supervision for Stylized Opinion Summarization}

\author{Hayate Iso \\
Megagon Labs\\
\small{\texttt{hayate@megagon.ai}} \And
Xiaolan Wang\thanks{~~Work done while at Megagon Labs.} \\
Meta Platforms, Inc.\\
\small{\texttt{xiaolan@meta.com}} \And
Yoshi Suhara$^*$ \\
NVIDIA\\
\small{\texttt{ysuhara@nvidia.com}}
}

\begin{document}
\maketitle
\begin{abstract}
Opinion summarization research has primarily focused on generating summaries reflecting important opinions from customer reviews without paying much attention to the writing style.
In this paper, we propose the stylized opinion summarization task, which aims to generate a summary of customer reviews in the desired (e.g., professional) writing style. 
To tackle the difficulty in collecting customer and professional review pairs, we develop a non-parallel training framework, Noisy Pairing and Partial Supervision (\napafont{Napa}\wine), which trains a stylized opinion summarization system from non-parallel customer and professional review sets. 
We create a benchmark {\sc ProSum} by collecting customer and professional reviews from Yelp and Michelin. Experimental results on {\sc ProSum} and FewSum demonstrate that our non-parallel training framework consistently improves both automatic and human evaluations, successfully building a stylized opinion summarization model that can generate professionally-written summaries from customer reviews.\footnote{The code is available at \url{https://github.com/megagonlabs/napa}}

\end{abstract}

\section{Introduction}
Opinion summarization, which focuses on automatically generating textual summaries from multiple customer reviews, has received increasing attention due to the rise of online review platforms.
Different from single-document summarization tasks (e.g., news summarization), which can easily collect a large amount of document-summary pairs, manually creating summaries from multiple reviews is expensive; it is not easy to collect large-scale training data for opinion summarization.
To address this challenge, existing studies build pseudo-reviews-summary pairs in a self-supervised fashion~\cite{chu2019meansum,amplayo-lapata-2020-unsupervised,suhara-etal-2020-opiniondigest,amplayo-etal-2021-aspect,iso-etal-2021-convex-aggregation} or use a small amount of reviews-summary pairs in a few-shot manner~\cite{brazinskas-etal-2020-shot,oved-levy-2021-pass,iso-etal-2022-comparative} to train opinion summarization models.

\begin{figure}
    \centering
    \includegraphics[width=\linewidth]{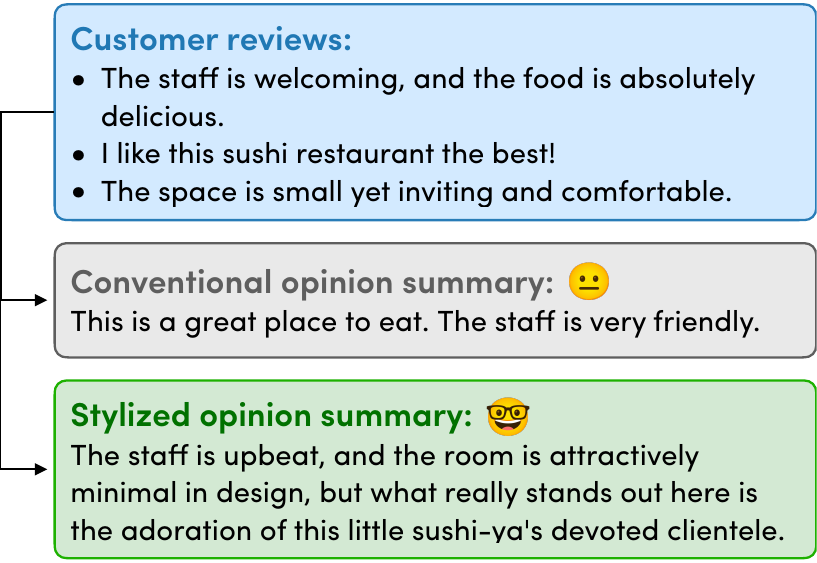}
    \caption{Comparison of conventional and stylized opinion summarization. Given multiple reviews as input, stylized opinion summarization aims to generate a summary in the desired writing style.%
    }
    \label{fig:task}
\end{figure}

\begin{figure*}[t]
     \centering
     \begin{subfigure}[b]{0.5\linewidth}
         \centering
         \includegraphics[height=3.5cm]{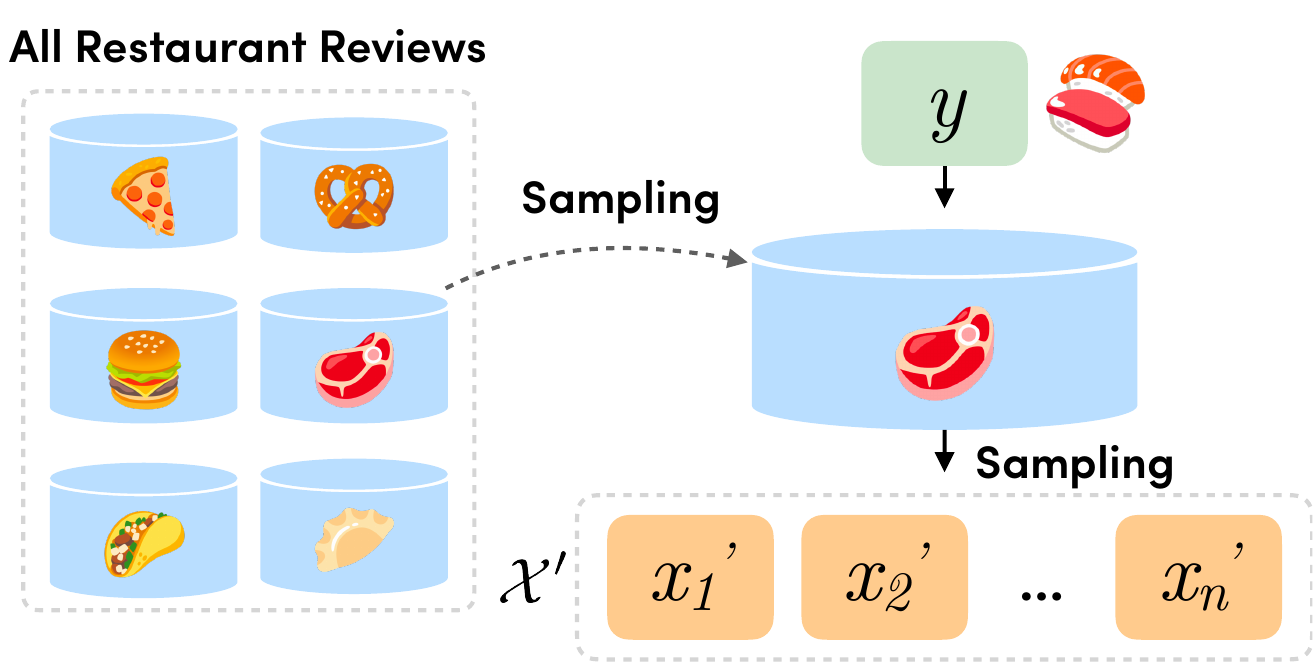}
         \caption{{\bf Noisy Pairing}: Given the candidate summary $y$, the pairs of noisy input reviews and output summary, $(\mathcal{X}', y)$, are built by retrieving the input reviews from a set of reviews from an arbitrary entity. This example retrieves the reviews from a steak restaurant given the professionally written summary of a sushi restaurant.}
         \label{fig:noisy_pairing}
     \end{subfigure}
     \hfill
     \begin{subfigure}[b]{0.48\linewidth}
         \centering
         \includegraphics[height=3.5cm]{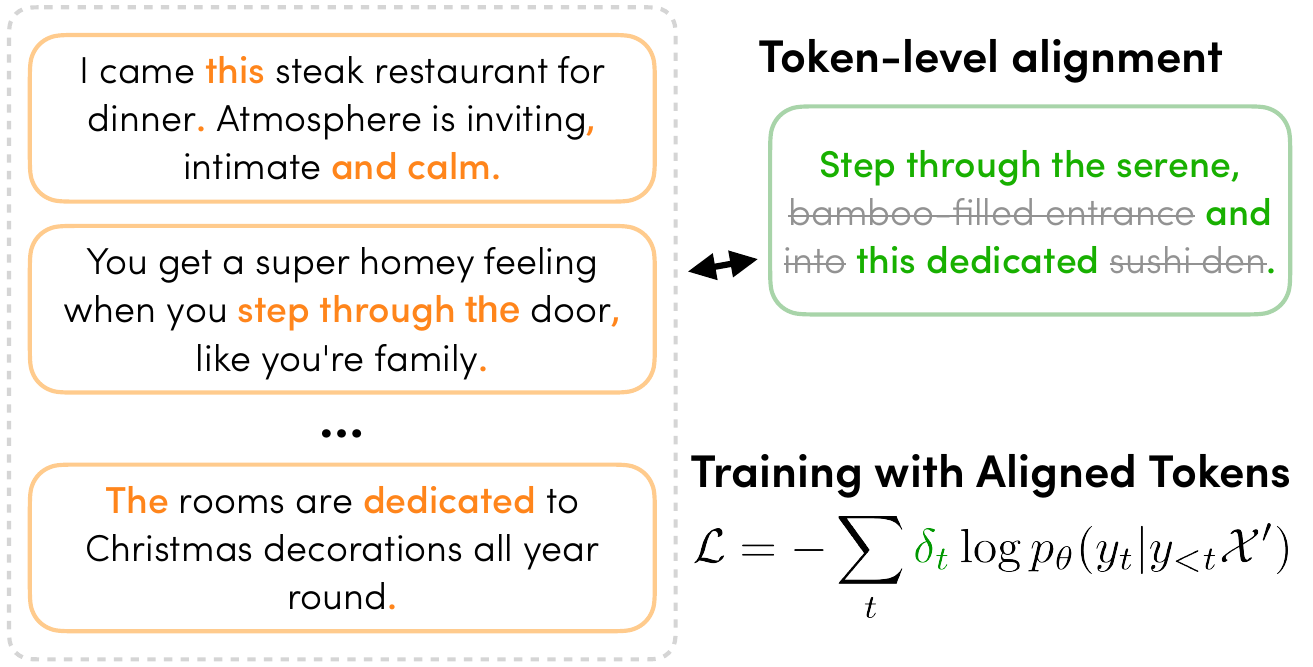}
         \caption{{\bf Partial Supervision}: After building a noisy input-output pair, we obtain the token-level alignment between the pair based on the word, stem, and synonym matching. Finally, we introduce indicator functions {\color{ggreen}$\delta_t$} into the standard negative log-loss function $\mathcal{L}$ to train using only aligned tokens, highlighted in {\color{ggreen}\textbf{green}}.}
         \label{fig:partial_supervision}
     \end{subfigure}
    \caption{Overview of our non-parallel training framework, Noisy Pairing and Partial Supervision.
    }
    \label{fig:overview}

    \end{figure*}

However, existing opinion summarization systems have focused on summarizing important opinions in reviews while not paying much attention to the writing style. They leverage customer reviews as pseudo summaries to train models, which generate summaries in the same writing style as the customer reviews 
as illustrated in Figure~\ref{fig:overview}.
On the other hand, professional reviews, such as Michelin Guide---a prestigious and popular restaurant guide, use a quite different writing style to describe the same type of information.

In this paper, we aim to fill this gap between customer and professional reviews by proposing a new branch of opinion summarization---{\em stylized opinion summarization}, where the goal is to generate a summary of opinions in the desired writing style.
Specifically, besides customer reviews, as the input to the conventional opinion summarization task, we use a few example summaries in the desired writing style as auxiliary information to guide the model in learning the writing style.
Since a few summaries in the desired writing style may not cover the same entities (e.g., restaurants) as the customer review set, the two review sets for the stylized opinion summarization task are non-parallel, which makes the task more challenging.\footnote{We will also evaluate the parallel setting later.}

To this end, we develop a non-parallel training
framework, \textit{Noisy Pairing and Partial Supervision} (\napafont{Napa}\wine), which builds a stylized opinion summarization model from {\em non-parallel} customer and professional review sets.
The core idea consists of two functions: {\em Noisy Pairing} (\S\ref{sub:noisy_pairing}) creates pseudo ``noisy'' reviews-summary pairs forcibly for each summary in the desired writing style by obtaining input reviews similar to the summary. Then, {\em Partial Supervision} (\S\ref{sub:partial_supervision}) trains a model with the collected noisy pairs by focusing on the sub-sequence of the summary that can be reproduced from the input reviews while not learning to hallucinate non-existing content. 
Figure~\ref{fig:overview} illustrates the two functions. In this example, for a professionally-written review of a sushi restaurant, Noisy Pairing finds reviews of a steak restaurant as noisy source reviews, which are then {\em partially} used by Partial Supervision to train a stylized opinion summarization model.

We also create and release a benchmark for stylized opinion summarization named \textsc{ProSum}, which consists of 700 paired Yelp reviews and Michelin point-of-views.
Experimental results on \textsc{ProSum} confirm that \napafont{Napa}\wine successfully generates summaries in the desired writing style in a non-parallel training setting, significantly better than models trained by self-supervision and existing non-parallel training methods.

We further performed additional experiments using existing supervised opinion summarization benchmarks, FewSum~\cite{brazinskas-etal-2020-shot}, 
in a non-parallel setting.
We observed that \napafont{Napa}\wine brings significant gains
over self-supervised systems and competitive performance with state-of-the-art supervised systems, indicating the generalizability of the proposed method. %

\section{The \textsc{ProSum} Corpus}
\begin{table*}[t]
    \centering
    \footnotesize
    \begin{tabular}{l|cc|cccc|ccc}
    \toprule
    & \multirow{2}{*}{Src len.} & \multirow{2}{*}{Tgt len.} & \multicolumn{4}{c}{\% of novel $n$-grams in gold summary} & \multicolumn{3}{c}{Extractive oracle}\\
    & & & Unigram & Bigram & Trigram & 4-gram & R1 & R2 & RL \\\midrule
    \textsc{ProSum}~\scriptsize{(ours)} & 1162.7 & 139.7 & 38.19 & 84.76 & 97.17 & 99.18 & 42.97 & 10.99 & 22.59\\\midrule
    Yelp~\scriptsize{\cite{brazinskas-etal-2020-shot}} & 453.3 & 58.02 & 31.71 & 83.02 & 95.53 & 98.35 & 47.79 & 15.28 & 25.84\\
    Amazon~\scriptsize{\cite{brazinskas-etal-2020-shot}} & 446.2 & 56.89 & 31.62 & 82.32 & 95.84 & 98.60 & 46.31 & 14.27 & 25.44\\
    \bottomrule
    \end{tabular}
    \caption{Statistics of \textsc{ProSum} and FewSum Yelp/Amazon benchmarks. \textsc{ProSum} has a longer source and target length compared to the FewSum benchmarks and offers more abstractive summaries with respect to the novel $n$-gram ratio. The source and target length is the number of BPE tokens per example using the BART tokenizer.
    }
    \label{tab:stats}
\end{table*}

\paragraph{Data Collection}
We build a stylized opinion summarization dataset, \textsc{ProSum}, which pairs customer reviews and professional reviews about the same restaurant, as we need customer reviews as the input and a professional review as the summary for evaluation purposes.

We first collected 700 professionally-written restaurant reviews from \url{guide.michelin.com}, a famous restaurant review site. Unlike crowd-sourced opinion summaries, these reviews are written by professional writers. Thus, they include more appealing expressions and attractive information than crowd-sourced summaries.
Then, we collected customer reviews from a popular customer review platform, \url{yelp.com}, by asking crowdsourced workers from Appen\footnote{\url{https://appen.com/}} to find the same restaurant for each of the restaurants we collected in the first step. We collected up to 5,000 customer reviews for each restaurant.

\paragraph{Filtering}
Since our main focus is to create a stylized opinion summarization benchmark and thousands of input reviews cannot be handled by most pre-trained language models, we filtered source customer reviews to reduce the number of input reviews to a size that can be handled by commonly used pre-trained language models. 

For each reviews-summary pair, we selected source Yelp reviews so that the coverage of the target Michelin review was maximized. Specifically, we used the sum of the ROUGE-1/2 Recall scores between the selected source Yelp reviews and the target Michelin review to measure the coverage. We incrementally added source reviews until the total length exceeded 1,024 words to maximize the coverage in a greedy manner. On average, 6.7 input reviews were selected for each pair. This selection step is to ensure the target Michelin summary 
can be created by source Yelp reviews.

Finally, we shuffled the selected source reviews to remove the selection order bias.
The final benchmark consists of 100/100/500 entities for the training/validation/test set.
Note that we keep parallel data (i.e., reviews-summary pairs) in {\sc ProSum} for evaluation and for training supervised models. For \napafont{Napa}\wine or other non-parallel training models, we  remove source reviews from the training set.

\paragraph{Statistics}
We summarize the \textsc{ProSum} dataset and compare it with existing opinion summarization datasets in Table~\ref{tab:stats}. We calculate novel $n$-grams in gold summaries to evaluate how abstractive/extractive \textsc{ProSum} is and the performance of the extractive oracle summaries from the source reviews.
We confirm that the \textsc{ProSum} is more abstractive than the existing benchmarks. The extractive oracle performance supports the feasibility of stylized opinion summarization in {\sc ProSum}.

\section{Self-supervised Opinion Summarization}
\label{sec:preliminary}
This section describes the standard self-supervised framework for conventional opinion summarization and then the pseudo-reviews-summary pair construction approach~\cite{elsahar-etal-2021-self}, which is also used as the pre-training method in \S\ref{sec:eval}.

Opinion summarization is a multi-document summarization problem that aims to generate a textual summary text $y$ that reflects the salient opinions given the set of reviews $\mathcal{X} = \{x_1, \dots, x_N\}$.
Due to the unavailability of a sufficient amount of reference summaries for training, a commonly used approach is to create a pseudo-reviews-summary training pair $(\tilde{\mathcal{X}}, \tilde{y})$ from a massive amount of reviews and trains an opinion summarization model $p_{\theta}$ using negative log-loss minimization,
\begin{align*}
    \mathcal{L} = -\log p_{\theta}(\tilde{y} | \tilde{\mathcal{X}}) = -\sum_t \log p_{\theta}(\tilde{y}_t | \tilde{y}_{<t}, \tilde{\mathcal{X}}).
\end{align*}

\paragraph{Pseudo reviews-summary pairs construction}

Let $\mathcal{R}_e$ denotes the set of reviews for specific entity $e$ such as a restaurant.
For each set of reviews $\mathcal{R}_e$, we treat a review in this set as a pseudo summary $\tilde{y} \in \mathcal{R}_e$ and then retrieve the relevant reviews to build a source set of reviews $\tilde{\mathcal{X}}$.
Concretely, given a pseudo summary $\tilde{y}$, retrieve the source set of $N$ reviews $\tilde{\mathcal{X}}$ by maximizing the sum of the similarity as follows:
\begin{align*}
    \tilde{\mathcal{X}} = \argmax_{\mathcal{X} \subset \mathcal{R}_e \setminus \{\tilde{y}\}, |\mathcal{X}| = N} \sum_{x \in \mathcal{X}}\text{sim}(x, \tilde{y}) , \label{eq:pairing}
\end{align*}
where similarity is measured by the cosine similarity of the TF-IDF vector.
This operation is applied to all reviews as pseudo summaries. Then the top-$K$ pseudo-reviews-summary pairs with the highest similarity scores $\sum_{x \in \tilde{X}} \text{sim}(x, \tilde{y})$ are retained as the final pseudo-training set $\{(\tilde{\mathcal{X}}_i, \tilde{y}_i)\}_{i=1}^K$.

\section{\napafont{Napa}\wine}
\label{sec:method}

Although pseudo-reviews-summary pairs creation has been one of the solid approaches for conventional opinion summarization, we cannot directly use it for stylized opinion summarization, as there are two sets of {\em non-parallel} reviews in different writing styles.

This section describes a non-parallel training framework for stylized opinion summarization, \textit{Noisy Pairing and Partial Supervision} (\napafont{Napa}\wine), which trains a summarization model from non-parallel customer and professional review sets.

\subsection{Noisy Pairing}
\label{sub:noisy_pairing}
Noisy Pairing expands the existing pseudo-reviews-summary construction approach to create “noisy” reviews-summary pairs for each summary in the desired writing style by obtaining input reviews similar to the summary.

To leverage the desired style of summary $y$ for the entity $e$, which is not paired with the set of reviews for the same entity $\mathcal{R}_e$, we first build the \textit{noisy} reviews-summary pairs.
Specifically, given the summary $y$ for entity $e$, we follow the pseudo data construction approach (\S\ref{sec:preliminary}) to construct the source set of reviews, but we retrieve the reviews from the \textit{different} entity $e' (\neq e)$ with the summary:
\begin{align*}
    \tilde{\mathcal{X}}' = \argmax_{\mathcal{X} \subset \mathcal{R}_{e'}, |\mathcal{X}| = N} \sum_{x \in \mathcal{X}}\text{sim}(x, y).
\end{align*}
For instance, given a summary of a sushi restaurant, we can use reviews of a steak restaurant to construct a noisy reviews-summary pair as illustrated in Figure~\ref{fig:overview}.
Then, using the similar approach used in the pseudo data construction, we obtain the final noisy training set  $\{(\tilde{\mathcal{X}}', y)\}$.
In particular, the top 10 noisy reviews-summary pairs of the highest similarity score are retained for each summary.

Note that this method could unintentionally select the review of the correct entity as input (i.e., $e' = e$), so in our experiments, we explicitly discarded the review of the entity used in summary to maintain the non-parallel setting.

\subsection{Partial Supervision}
\label{sub:partial_supervision}
With the noisy pairing method described above, we can build noisy reviews-summary pairs $\{(\tilde{\mathcal{X}}', y)\}$, but obviously, a model trained with these pairs will generate unfaithful summaries.
However, even in such noisy reviews-summary pairs, there would be sub-sequences of the summary $y$ that could be generated from noisy input reviews $\tilde{\mathcal{X}}'$.

To implement this intuition into the training, we first compute the \textit{token-level alignment} between a noisy set of reviews $\tilde{\mathcal{X}}'$ and summary $y$, and then introduce the indicator function $\delta_t$ inside of the standard log-loss function to ignore the unaligned tokens during the training:
\begin{align*}
    \mathcal{L}' = -\sum_t \delta_t \log p_{\theta}(y_t | y_{<t}, \tilde{\mathcal{X}}'),
\end{align*}
where the %
alignment function $\delta_t$ will be 1 if the token $y_t$ is aligned with the noisy source reviews $\mathcal{X}$ and otherwise 0 as illustrated in Figure~\ref{fig:partial_supervision}.
This allows for using aligned words, such as the style and expressions used in the summary, as a training signal without increasing the likelihood of hallucinated words.

For the alignment function, we use word-level matching between the source and target reviews. Since professional writers have a rich vocabulary, which contains words that rarely appear in customer reviews, we implement word stem matching and synonym matching (e.g., serene $\sim$ calm) to increase the coverage in Partial Supervision. We discuss the design choice of the alignment function in \S\ref{sub:alignment}. 

\section{Evaluation}\label{sec:eval}

We use \textsc{ProSum} and an existing opinion summarization benchmark FewSum~\cite{brazinskas-etal-2020-shot} to verify the effectiveness and generalizability of \napafont{Napa}\wine.
For FewSum, we discarded the source reviews from the training dataset to convert FewSum into a stylized opinion summarization benchmark (i.e., in the non-parallel setting).

\subsection{Settings}
\paragraph{Training Data}

For non-parallel training, we first pre-train a self-supervised opinion summarization model using pseudo-reviews-summary pairs (\S\ref{sec:preliminary}). Then, we fine-tune it using noisy reviews-summary pairs using \napafont{Napa}\wine (\S\ref{sec:method}). Therefore, we need two sets of pseudo-reviews-summary pairs for self-supervised pre-training and noisy reviews-summary pairs for \napafont{Napa}\wine.

As {\sc ProSum} does not contain customer reviews for training, we use the Yelp review dataset\footnote{\url{https://www.yelp.com/dataset}}, which has 7M reviews for 150k entities, to collect reviews-summary pairs for \textsc{ProSum} dataset.
We discarded all the entities used in the Michelin reviews in {\sc ProSum} to avoid unintentionally selecting the same entity for Noisy Pairing. Then, we excluded entities that do not satisfy the following criteria: (1) in either the \texttt{restaurant} or \texttt{food} category; (2) the rating is higher than 4.0/5.0 on average. Then, we filtered reviews with 5-star ratings.
Finally, we discarded entities that have less than ten reviews. After this pre-processing, we built 100k pseudo-reviews-summary pairs and 1k noisy reviews-summary pairs for self-supervised pre-training and \napafont{Napa}\wine, respectively.
The pre-processing method for the FewSum dataset is described in Appendix.

\paragraph{Model}
We instantiate our summarization models using the Transformer model~\cite{vaswani2017transformer} initialized with the \texttt{BART-large} checkpoint~\cite{lewis-etal-2020-bart} in the \texttt{transformers} library~\cite{wolf-etal-2020-transformers}.
We used AdamW optimizer~\cite{loshchilov2019decoupled} with a linear scheduler and warmup, whose initial learning rate is set to 1e-5, and label smoothing~\cite{szegedy2016rethinking} with a smoothing factor of 0.1.
We tested three configurations: (1) the full version, (2) without Partial Supervision, and (3) without Noisy Paring and Partial Supervision---the self-supervised base model trained only using pseudo-review-summary pairs.

\subsection{Baselines}
For the main experiment on {\sc ProSum}, we compared the state-of-the-art opinion summarization system (BiMeanVAE) and two text-style transfer models (Pipeline and Multitask). 
We also evaluated the upper-bound performance of \napafont{Napa}\wine by using the {\em parallel} training dataset, where the customer and professionally written reviews for the same entity are correctly paired (Supervised upperbound). 
For the FewSum dataset, we compared various opinion summarization models, including self-supervised models and supervised models that use parallel training data, to verify the performance of our non-parallel training framework. The details can be found in Appendix.

\paragraph{BiMeanVAE:}
BiMeanVAE~\cite{iso-etal-2021-convex-aggregation} is a self-supervised opinion summarization model based on a variational autoencoder. 
We further fine-tune this model using Michelin reviews to generate summaries with the desired style.

\paragraph{Pipeline:}
We combine a self-supervised opinion summarization model and text style transfer model to build a two-stage pipeline. For the self-supervised model, we use the same self-supervised base model as \napafont{Napa}\wine.
For the text style transfer model, we use STRAP~\cite{krishna-etal-2020-reformulating}, which uses inverse paraphrasing to perform text style transfer using Yelp and Michelin reviews in the non-parallel setting.

\paragraph{Multitask: }
We use a multi-task learning framework, TitleStylist~\cite{jin-etal-2020-hooks}, which combines summarization and denoising autoencoder objectives to train a summarization model that generates summaries in the desired writing style. 
In the experiment, we use Yelp pseudo-reviews-summary pairs (Michelin reviews) for the summarization (denoising) objective. %

\begin{table*}[t]
    \centering
    \resizebox{\textwidth}{!}{
    \begin{tabular}{l|cccccc}
    \toprule
        & \multicolumn{5}{c}{\textsc{ProSum}}\\
        & R1 & R2 & RL & BS & Consistency & Relevance\\\midrule
        \textbf{Non-parallel baselines}\\
        \quad Multitask~\footnotesize{\cite{jin-etal-2020-hooks}} & 23.78 & 1.85 & 15.81 & 80.92& 95.01 & 89.84\\
        \quad Pipeline~\footnotesize{\cite{krishna-etal-2020-reformulating}} & 27.19 & 2.69 & 16.76 & 82.88 & 96.69 & 91.99\\
        \quad BiMeanVAE~\footnotesize{\cite{iso-etal-2021-convex-aggregation}} & 28.15 & 3.49 & 18.68 & 83.10 & 96.83 & 91.98\\\midrule
        \napafont{Napa}\wine\\
        \qquad Full version & \textbf{33.54} & \textbf{4.95} & \textbf{20.67} & \textbf{84.77} & 96.86 & \textbf{92.48} \\
        \qquad w/o Partial Supervision & 31.64 & 3.96 & 18.90 & 84.15 & 96.09 & 91.80 \\
        \qquad w/o Noisy Paring and Partial Supervision & 28.19 & 3.43 & 17.60 & 83.49 & \textbf{96.88} & 91.92 \\
        \midrule
        \textbf{Supervised upperbound} & 34.50 & 5.70 & 20.64 & 84.96 & 97.23 & 92.96 \\
    \bottomrule
    \end{tabular}
    }
    \caption{
    Experimental results on the \textsc{ProSum} dataset. R1/2/L and BS denote the F1 scores of ROUGE-1/2/L and BERTScore.
    \napafont{Napa}\wine gives substantial improvements over the baselines. We also confirm that Partial Supervision successfully alleviates the consistency degradation caused by Noisy Pairing.}
    \label{tab:main}
\end{table*}

\begin{table*}[ht]
    \centering
    \resizebox{0.9\textwidth}{!}{
    \begin{tabular}{l|ccc|ccc}
    \toprule
         & \multicolumn{3}{c|}{\textsc{Yelp}} & \multicolumn{3}{c}{\textsc{Amazon}} \\
         & R1 & R2 & RL & R1 & R2 & RL\\\midrule
         \textbf{Self-supervised baselines} & & & & &  \\
         \quad MeanSum~\footnotesize{\cite{chu2019meansum}} & 27.50 & 3.54 & 16.09 & 26.63 & 4.89 & 17.11 \\
         \quad CopyCat~\footnotesize{\cite{brazinskas-etal-2020-unsupervised}} & 28.12 & 5.89 & 18.32 & 27.85 & 4.77 & 18.86 \\
         \midrule
         \textbf{Supervised baselines} -- Parallel training & & & & \\
         \quad FewSum~\footnotesize{\cite{brazinskas-etal-2020-shot}} & 37.29 & 9.92 & 22.76 & 33.56 & 7.16 & 24.49\\
         \quad PASS~\footnotesize{\cite{oved-levy-2021-pass}} & 36.91 & 8.12 & 23.09 & 37.43 & 8.02 & 23.34\\
         \quad AdaSum~\footnotesize{\cite{brazinskas-etal-2022-efficient}} & 38.82 & 11.75 & 25.14 & 39.78 & 10.80 & 25.55\\
         \quad BART~\footnotesize{(our implementation)} & 39.69  & 11.63 & 25.48 & 39.05 & 10.08 & 24.29\\\midrule
         \napafont{Napa}\wine -- Non-parallel training & & & & \\
         \quad Full version & \textbf{38.59} & \textbf{11.23} & \textbf{25.29} & \textbf{36.21} & \textbf{9.18} & \textbf{23.60} \\
         \quad w/o Partial Supervision & 37.41 & 10.51 & 24.18 & 35.30 & 7.45 & 21.92 \\
         \quad w/o Noisy Pairing and Partial Supervision & 33.39 & 7.64 & 20.67 & 30.18 & 5.24 & 19.70\\         
         \bottomrule
    \end{tabular}
    }
    \caption{
    Experimental results on the FewSum dataset~\cite{brazinskas-etal-2020-shot}. \napafont{Napa}\wine shows substantial improvements over the self-supervised baselines. Note that the supervised baseline models were fine-tuned on the parallel training data (i.e., annotated reviews-summary pairs), while \napafont{Napa}\wine models were trained in the non-parallel setting.}
    \label{tab:sub}
\end{table*}

\subsection{Automatic Evaluation}

We use the F1 scores of ROUGE-1/2/L~\cite{lin-2004-rouge}\footnote{\url{https://github.com/Diego999/py-rouge}} and BERTScore~\cite{Zhang*2020BERTScore:}\footnote{\url{https://github.com/Tiiiger/bert\_score}} for reference-based automatic evaluation.
Additionally, we calculate the CTC score~\cite{deng-etal-2021-compression} to evaluate the consistency and relevance of the generated summaries. 
The consistency score is measured by the alignment between the source reviews and the generated summary based on the contextual embedding similarity; the relevance score is measured by the alignment between the generated summary and the reference summary multiplied by the consistency score.
The contextual embeddings are obtained from the \texttt{roberta-large} model.

\paragraph{ProSum}

Table~\ref{tab:main} shows the main experimental results on \textsc{ProSum}.
The self-supervised model (i.e., \napafont{Napa}\wine w/o Noisy Pairing and Partial Supervision) outperforms all the non-parallel baseline systems.
The comparison shows that Pipeline, which combines the self-supervised model and STRAP, degrades the summarization quality. The result indicates that it is not easy to achieve stylized opinion summarization by simply combining a summarization model and a text style transfer model. 

\napafont{Napa}\wine w/o Partial Supervision improves the summarization quality against the self-supervised model while causing degradation in consistency between generated summaries and the source reviews. This degradation is expected, as Noisy Pairing creates pseudo-reviews-summary by sampling reviews from a different entity, only considering the similarity against the pseudo-summary. We will discuss this point in detail in \S\ref{sub:why}.

\napafont{Napa}\wine substantially outperforms the baselines for summarization quality and relevance while maintaining the same level of consistency as the best self-supervised model. This confirms that Partial Supervision successfully alleviates the consistency degradation caused by Noisy Pairing. 

The experimental results demonstrate that both Noisy Pairing and Partial Supervision are essential to building a robust stylized opinion summarization model, allowing the model to take advantage of useful signals in the noisy reviews-summary pairs.

\paragraph{FewSum}
The experimental results on FewSum in the non-parallel setting shown in Table~\ref{tab:sub} also observe the substantial improvements by \napafont{Napa}\wine over the self-supervised systems.
\napafont{Napa}\wine shows competitive performance against state-of-the-art supervised systems, which use parallel training data for training. 
The results further confirm that providing a small number of reference summaries in the desired writing style, even if they are not paired with input reviews, can help \napafont{Napa}\wine train a solid summarization model for stylized opinion summarization.

\subsection{Human Evaluation}

We conducted human evaluations to compare the performance of our model (\napafont{Napa}\wine) with three baselines: Self-supervision, Pipeline, and \napafont{Napa}\wine without Partial Supervision (PS) on \textsc{ProSum} with respect to the fluency, relevance, and attractiveness of the generated summary. We asked human annotators recruited from Appen to rate generated summaries on a 4-point Likert scale for each evaluation metric. We describe more details of the human evaluation in Appendix.

Our findings from the results shown in Figure~\ref{fig:human_eval} are: (1) using professionally-written summaries for training allows the model to generate more fluent and attractive summaries than other baselines (\napafont{Napa}\wine and \napafont{Napa}\wine w/o PS vs. Self-supervision and Pipeline); (2) \napafont{Napa}\wine without Partial Supervision tends to generate more irrelevant summaries (\napafont{Napa}\wine vs. \napafont{Napa}\wine w/o PS).
Overall, our results demonstrate the importance of using professionally-written summaries for training to improve the fluency and attractiveness of generated summaries and the need for Partial Supervision to ensure the relevance of generated summaries.

\begin{figure}[t]
    \centering
    \includegraphics[width=\linewidth]{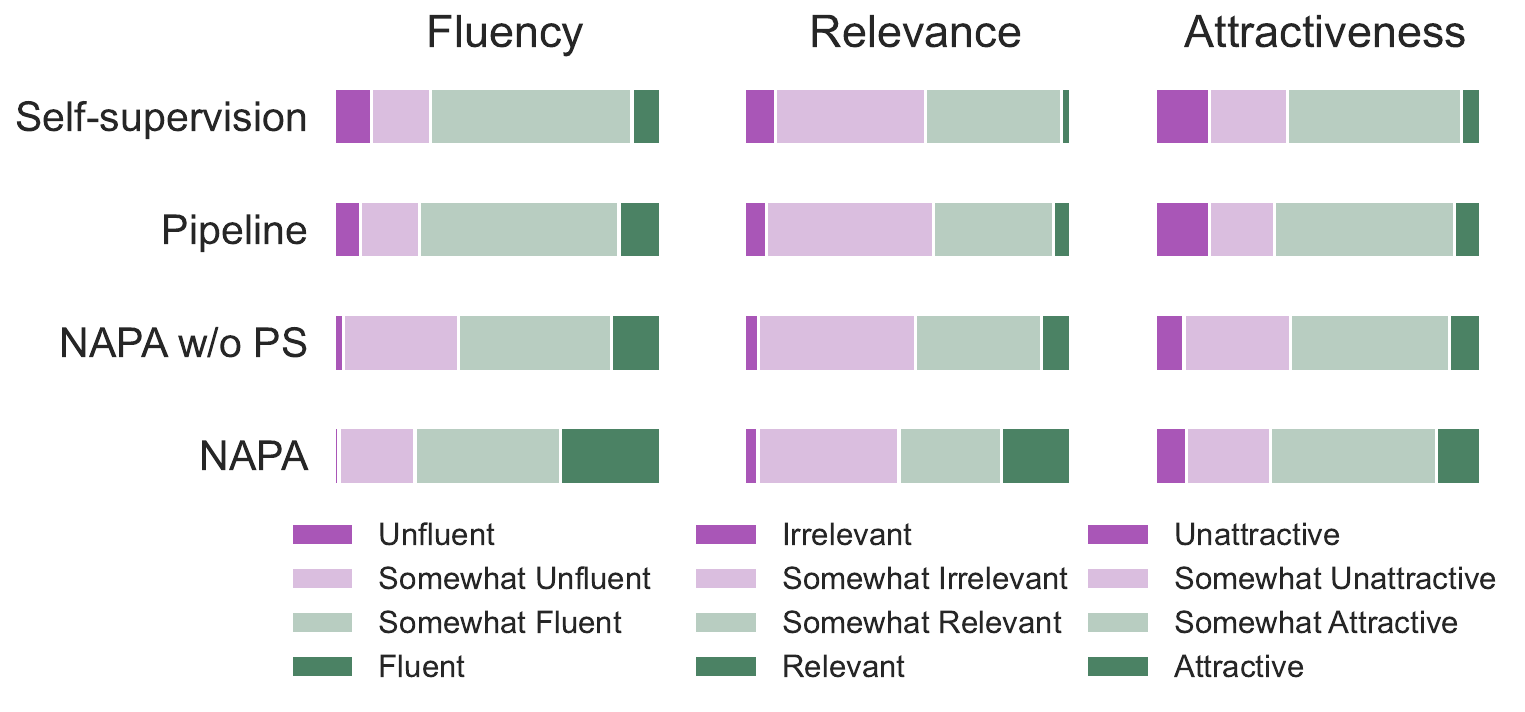}
    \caption{Human evaluations of the fluency, relevance, and attractiveness on \textsc{ProSum}.}
    \label{fig:human_eval}
\end{figure}

\section{Analysis}
\subsection{Importance of Partial Supervision}
\label{sub:why}

The experimental results in Tables~\ref{tab:main} and \ref{tab:sub} show that \napafont{Napa}\wine without Partial Supervision---just using noisy reviews-summary pairs---demonstrates solid performance for reference-based automatic evaluation metrics. This is a little bit counterintuitive, and this can be attributed to the positive effect of early stopping against noisy training data~\cite{arpit2017closer,li2020gradient}.
To analyze this point, we conducted an additional experiment by training \napafont{Napa}\wine with and without Partial Supervision for more training epochs. 

\begin{figure}[t]
    \centering
    \includegraphics[width=\linewidth]{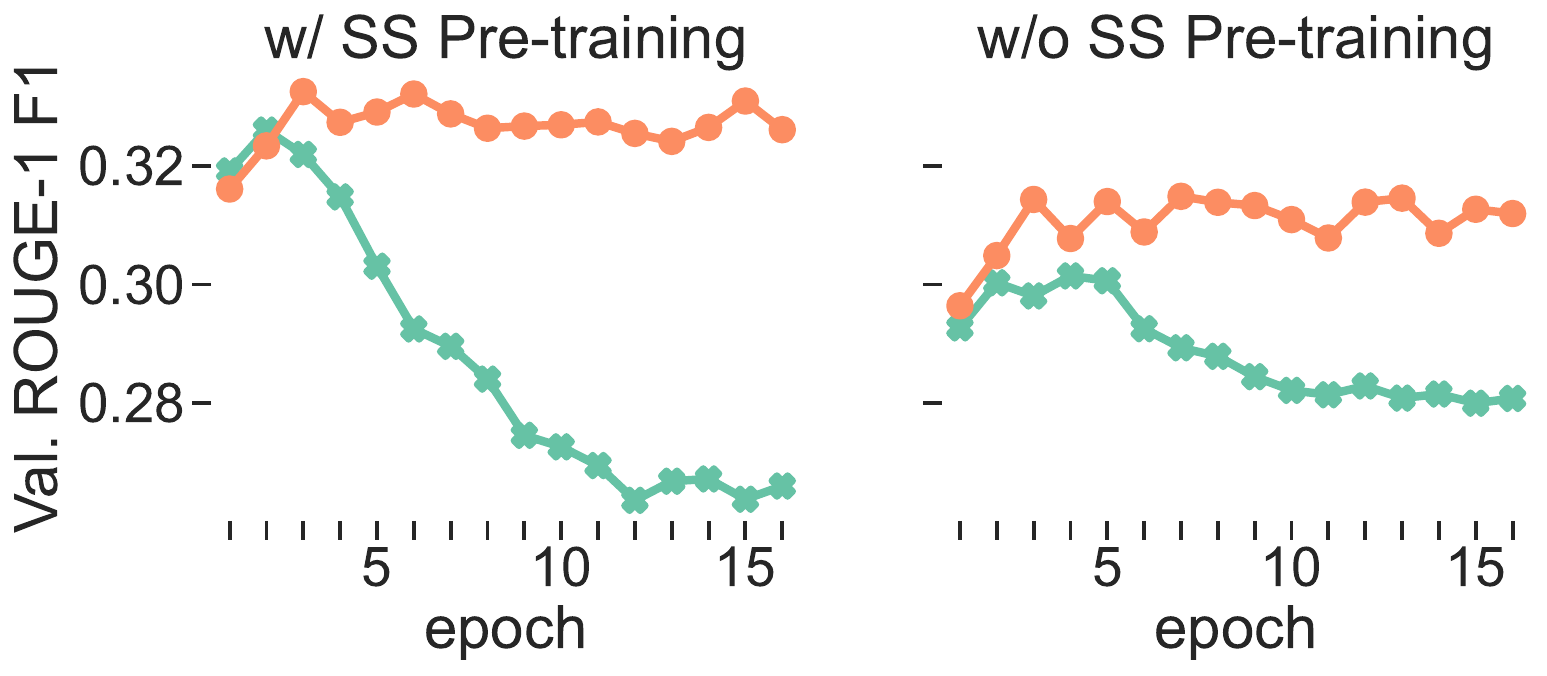}
    \caption{ROUGE-1 F1 score on validation set of \textsc{ProSum} at different training stages. The \textbf{\textcolor{oorange}{orange line}} denotes the model trained \textit{with} partial supervision (\S\ref{sub:partial_supervision}), and the \textbf{\textcolor{ggreen}{green line}} denotes the model trained \textit{without} partial supervision.
    }
    \label{fig:training}
\end{figure}

Figure~\ref{fig:training} shows the ROUGE-1 F1 score on the validation set of \textsc{ProSum} at different training epochs of the \napafont{Napa}\wine model trained \textit{with} or \textit{without} Partial Supervision (\textbf{\textcolor{oorange}{orange line}} and \textbf{\textcolor{ggreen}{green line}}).
As shown in the figure, we find that in the very early stages of training, both the models improve the ROUGE scores. In the later stage, \napafont{Napa}\wine {\em without} Partial Supervision (\textbf{\textcolor{ggreen}{green line}}) shows continuous degradation, while \napafont{Napa}\wine {\em with} Partial Supervision (\textbf{\textcolor{oorange}{orange line}}) shows robust performance consistently over the entire training process.

This observation is aligned with the literature on noisy supervision, which shows that over-parametrized models learn simple patterns in the early stages of training and then memorize noise~\cite{arpit2017closer}.
On the other hand, it is also known that early stopping is not sufficient under labeling noise~\cite{ishida2020flooding}.
We observed that \napafont{Napa}\wine \textit{without} Partial Supervision generated summaries that were less consistent with the source reviews (Table~\ref{tab:main}) and contained more hallucinations, as described in Appendix.
The results support the importance of Partial Supervision for improving the robustness of the stylized opinion summarization model in non-parallel training.

\begin{figure}[t]
    \centering
    \includegraphics[width=\linewidth]{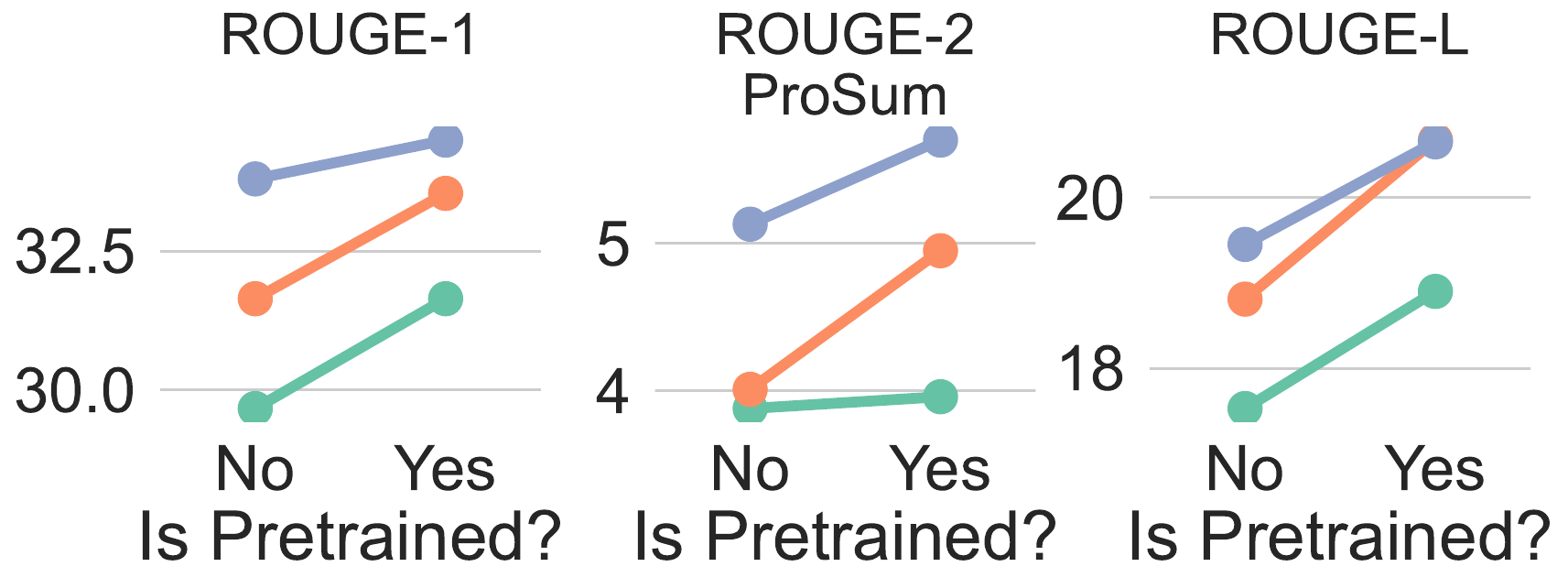}
    \caption{Comparison of summarization quality with and without pre-training. The \textbf{\textcolor{bblue}{blue line}} denotes the model trained in a supervised setting, \textbf{\textcolor{oorange}{orange line}} denotes the model trained \textit{with} partial supervision and \textbf{\textcolor{ggreen}{green line}} denotes the model trained \textit{without} partial supervision.
    }\label{fig:ablation}
\end{figure}

\begin{table*}[ht]
    \centering
    \resizebox{\textwidth}{!}{
    \begin{tabular}{l|cccc|cccc}
        \toprule
        & \multicolumn{4}{c|}{Reference based metrics} & \multicolumn{4}{c}{Novel $n$-gram ratios}\\
        & R1 & R2 & RL & BS & Unigram & Bigram & Trigram & Four-gram \\\midrule
        \napafont{Napa}\wine & & & & &\\
        \quad No Partial Supervision ($\delta_t = 1$ for all $t$) & 31.64 & 3.96 & 18.90 & 84.15 & 31.52 & 80.38 & 96.54 & 99.23\\
        \qquad + word match & 32.88 & 4.77 & 19.98 & 84.50 & 12.78 & 64.10 & 91.63 & 97.69\\
        \qquad + word or stem match & 32.49 & 4.82 & 20.03 & 84.45 & 13.23 & 66.60 & 92.27 & 97.94 \\
        \qquad + word or stem or synonym match & 33.54 & 4.95 & 20.67 & 84.77 & 15.54 & 67.19 & 92.24 & 97.75 \\
        \midrule
        \textbf{Supervised upperbound} & 34.50 & 5.70 & 20.65 & 84.96 & 14.59 & 58.84 & 83.20 & 91.38\\
        \bottomrule
    \end{tabular}
    }
    \caption{Comparison of summaries generated with different alignment criteria; + word match is the strictest alignment criterion; adding + stem and + synonym match allows for more relaxed alignment criteria allowing more words to be used for training. As the alignment criteria are relaxed, more novel $n$-grams can be generated.}
    \label{tab:choice}
\end{table*}

\subsection{Pre-training with Self-supervision}\label{sub:pretrain}
As we observe that the self-supervised baseline (i.e., \napafont{Napa}\wine w/o Noisy Pairing and Partial Supervision) shows solid performance in Table~\ref{tab:main} and better performance than the other self-supervised baselines in Table~\ref{tab:sub}, we further investigated the effectiveness of the pre-training using pseudo-reviews-summary pairs (Self-supervision in \S\ref{sec:preliminary}) in the non-parallel training.
We conducted ablation studies for the model trained \textit{with} Partial Supervision (\textbf{\textcolor{oorange}{orange line}}), \textit{without} Partial Supervision (\textbf{\textcolor{ggreen}{green line}}), and supervised setting (\textbf{\textcolor{bblue}{blue line}}).%

As shown in Figure~\ref{fig:ablation}, pre-training with self-supervision in all the settings helps improve summarization quality. The effect of pre-training is the most remarkable in the non-parallel settings (\textbf{\textcolor{oorange}{orange line}} and \textbf{\textcolor{ggreen}{green line}}).
This indicates that while non-parallel training helps learn the desired writing style for summary generation, it is difficult to determine what content to include in the summary only from the noisy-reviews-summary pairs.
Therefore, we experimentally confirm the effectiveness of self-supervised pre-training for stylized opinion summarization; self-supervision pre-training teaches the model the basics of how to summarize the content, and non-parallel training introduces the model to write in the desired style. The same analysis on the FewSum dataset can be found in Appendix.

\subsection{Choice of Token Alignment}\label{sub:alignment}

As discussed in \S\ref{sub:partial_supervision}, the token alignment function should be carefully chosen to appropriately align customer and professional reviews with different vocabularies. For example, the exact word match should naively disregard semantically similar words (e.g., serene and calm). 
Thus, we further performed a comparative analysis of the token alignment function. We compared \napafont{Napa}\wine with different variants of Partial Supervision that use: (1) exact word matching, (2) stem matching, and (3) synonym matching.

As shown in Table~\ref{tab:choice}, No Partial Supervision (first row) generates too many novel $n$-grams, indicating significant hallucinations; it shows the worst summarization performance.
We confirm that the model tends to generate more novel $n$-grams when the alignment criterion is relaxed and also improves summarization performance, suggesting that the stem and synonym matching functions can successfully consider semantically similar tokens to incorporate into training without degradaging the summarization performance.

\section{Related Work}
\paragraph{Opinion Summarization}
Due to the challenges in collecting training data,
many studies have developed unsupervised solutions for opinion summarization systems~\cite{chu2019meansum,amplayo-lapata-2020-unsupervised,suhara-etal-2020-opiniondigest,iso-etal-2021-convex-aggregation,basu-roy-chowdhury-etal-2022-unsupervised}. Recent studies have explored few-shot learning approaches that utilize a small number of review-summary pairs for training~\cite{brazinskas-etal-2020-shot,oved-levy-2021-pass,iso-etal-2022-comparative}.

Our technique falls in the middle of these two approaches, as we do not use annotated reviews-summary pairs for training while using a large number of customer reviews and a small number of professional reviews as auxiliary supervision signals. %

\paragraph{Text Style Transfer}
Text style transfer is a technique %
to rewrite the input text into the desired style~\cite{mcdonald-pustejovsky-1985-computational}.
The primary approach for text style transfer 
is \textit{sentence-level}, which is used as our baselines (Pipeline~\cite{krishna-etal-2020-reformulating} and Multitask~\cite{jin-etal-2020-hooks}).

Based on the observation that both Pipeline and Multitask do not perform well for the stylized opinion summarization task (in Table~\ref{tab:main}), we confirm that applying sentence-level style transfer cannot offer high-quality stylized opinion summarization and it requires \textit{paragraph-level} text style transfer, which needs further exploration~\cite{jin-etal-2022-deep}. 

\paragraph{Noisy Supervision}
Learning statistical models under labeling noise is a classic challenge in machine learning~\cite{angluin1988learning,natarajan2013learning} and is an active research field because of the increasing availability of noisy data~\cite{han2020survey,song2022learning}. 
Among the major approaches for noisy supervision, the loss adjustment approach is widely used in the NLP community, as it can be coupled with any type of commonly used Transformer-based language models~\cite{devlin-etal-2019-bert,brown2020gpt3}

In text generation, previous studies have attempted to improve the model faithfulness by treating hallucinated summaries as noisy supervision~\cite{kang-hashimoto-2020-improved,fu-etal-2020-partially,goyal-etal-2022-training}.
Our study is different from the line of work in the sense that we combine noisy-reviews-summary pairs and noisy supervision to develop a non-parallel training framework for stylized opinion summarization.

\section{Conclusions}
This paper proposes stylized opinion summarization, which aims to summarize opinions of input reviews in the desired writing style.
As parallel reviews-summary pairs are difficult to obtain, we develop a non-parallel training framework named Noisy Pairing and Partial Supervision (\napafont{Napa}\wine); it creates noisy reviews-summary pairs and then trains a summarization model by focusing on the sub-sequence of the summary that can be reproduced from the input reviews.
Experimental results on a newly created benchmark {\sc ProSum} and an existing opinion summarization benchmark FewSum demonstrate that our non-parallel training framework substantially outperforms self-supervised and text-style transfer baselines while competitively performing well against supervised models that use parallel training data.

\section{Limitations}
We do not see any ethical issues, but we would like to mention some limitations.
This study investigates the use of a limited number of unpaired desired summaries during training. We employ partial supervision to reduce the risk of hallucination, but there is still a potential to generate unfaithful summaries. Thus, the model may generate inconsistent opinions with the source reviews.
There is also a trade-off between the quality and diversity of our token-level alignment method. We decided to use exact, stem, and synonym-based matching, but these methods may introduce alignment errors, leading to noisier training.
For the annotation tasks, we paid \$0.96 for each summary for the crowd workers on Appen. The estimated hourly wage on the platform is \$13.48 per hour.
For the summary evaluation, we only used token-level matching metrics, unlike LLM-as-a-judge~\cite{liu-etal-2023-g,wu-etal-2024-less}.

\bibliography{custom}

\end{document}


\maketitle
\appendix

\section{Additional Experimental Details}

\subsection{Dataset splits}
We show the details of dataset splits in Table~\ref{tab:splits}. Note that we eliminate the source reviews for training to ensure the non-parallel setting. We only utilized the paired dataset to build the supervised upperbound model.
\begin{table}[t]
    \small
    \centering
    \begin{tabular}{c|ccc}
    \toprule
    & Train & Dev & Test \\\midrule
    \textsc{ProSum} & 100 & 100 & 500\\
    Yelp & 30 & 30 & 40\\
    Amazon & 28 & 12 & 20 \\
    \bottomrule
    \end{tabular}
    \caption{Details of dataset splits. Note that we eliminate the source reviews for training to ensure the non-parallel setting.}
    \label{tab:splits}
\end{table}

\subsection{Pre-processing decision on FewSum}
For the Yelp dataset, we used reviews provided in the Yelp Open Dataset.~\footnote{\url{https://www.yelp.com/dataset}} For the Amazon dataset, we used reviews in the Amazon product review dataset~\cite{he2016ups}. We specifically select 4 categories: \textit{Electronics; Clothing, Shoes and Jewelry, Home and Kitchen; Health and Personal Care}. Both datasets are available for academic purposes.

We first filter out the reviews shorter than 40 words and longer than 70 words and then remove the non-English reviews using the language identifier model implemented in \texttt{fasttext}~\cite{joulin-etal-2017-bag}.
Finally, we build the same approach to build pseudo and noisy pairs.

\subsection{Baselines on FewSum}
\begin{itemize}
    \item \textbf{MeanSum}~\cite{chu2019meansum}: the unsupervised single entity opinion summarization models based on autoencoders. It generates summaries from the averaged latent representations of reviews.
    \item \textbf{CopyCat}~\cite{brazinskas-etal-2020-unsupervised}: a single entity opinion summarization solution based on variational autoencoder models trained with leave-one-out objectives.
    \item \textbf{FewSum}~\cite{brazinskas-etal-2020-shot}: an extension of CopyCat model fine-tuned on FewSum dataset.
    \item \textbf{PASS}~\cite{oved-levy-2021-pass}: Fine-tuned transformer models initialized with T5 checkpoint~\cite{JMLR:v21:20-074} on FewSum dataset and LkO perturbations to select the subset of the representative input reviews to generate summaries.
    \item \textbf{AdaSum}~\cite{brazinskas-etal-2022-efficient}: Fine-tuned BART models on FewSum dataset with Adapter-tuning~\cite{pmlr-v97-houlsby19a} for parameter-efficient adaptation.
\end{itemize}

\subsection{Training details}
Major hyper-parameters for training models are reported in Table \ref{tab:hyperparam} following the "Show-You-Work" style suggested by \citet{dodge-etal-2019-show}.

\begin{table*}[ht]
    \centering
    \small
    \begin{tabular}{cc}
        \toprule
       \textbf{Computing infrastructure} & NVIDIA A100\\
       \midrule
       \textbf{Pre-training duration} & 24h \\
       \midrule
       \textbf{Fine-tuning duration} & 2h \\
       \midrule
       \textbf{Search strategy} & Manual tuning \\\midrule
       \textbf{Model implementation} & \url{[MASK]}\\
       \midrule
       \textbf{Model checkpoint} & \url{[MASK]}\\
       \bottomrule
    \end{tabular}

    \vspace{3mm}\begin{tabular}{ccc}
    \toprule
    \textbf{Hyperparameter} & \textbf{Search space} & \textbf{Best assignment} \\
    \midrule
    \# of self-supervision steps & 100,000 & 100,000\\ 
    \midrule
    \# of fine-tuning steps & 2,000 & 2,000\\ 
    \midrule 
    batch size & 8 & 8\\
    \midrule
    initial checkpoint & \texttt{facebook/bart-large} & \texttt{facebook/bart-large} \\
    \midrule
    label-smoothing~\tiny{\cite{szegedy2016rethinking}} & \emph{choice}[0.0, 0.1] & 0.1 \\
    \midrule
    learning rate scheduler & linear schedule with warmup & linear schedule with warmup\\
    \midrule
    warmup steps & 1,000 & 1,000 \\
    \midrule
    learning rate optimizer & AdamW~\tiny{\cite{loshchilov2019decoupled}} & AdamW~\tiny{\cite{loshchilov2019decoupled}}\\
    \midrule
    AdamW $\beta_1$ & 0.9 & 0.9\\
    \midrule
    AdamW $\beta_2$ & 0.999 & 0.999\\
    \midrule
    learning rate & 1e-5 & 1e-5 \\
    \midrule
    weight decay & \emph{choice}[0.0, 1e-3, 1e-2] & 1e-3 \\
    \midrule
    gradient clipping & 1.0 & 1.0 \\
    \bottomrule
    \end{tabular}
    \caption{\napafont{Napa}\wine search space and the best assignments.}
    \label{tab:hyperparam}
\end{table*}

\section{More Analysis}
\subsection{Manual evaluation of the \textsc{ProSum}}
Unlike existing opinion summarization datasets created by crowd workers, such as \textsc{FewSum}~\cite{brazinskas-etal-2020-shot}, the \textsc{ProSum} dataset is automatically created. Therefore, it is possible that the output summary may contain some content that cannot be recovered from the input reviews. Thus, we manually evaluated the quality of the \textsc{ProSum} dataset.

Specifically, we extracted noun phrases from the input reviews and output Michelin's point-of-view. Then, we evaluated how many noun phrases in the output summaries were contained in the input reviews. To make the evaluation more efficient, we used \texttt{sentence-transformers}~\url{https://github.com/UKPLab/sentence-transformers}~\cite{reimers-gurevych-2019-sentence} to extract the five noun phrases from the input reviews that were most similar to the noun phrases in the output summary.

We performed the evaluation on 20 randomly sampled pairs and found that 87.65\% of the noun phrases were also included in the input reviews. This result indicates that the majority of the facts were properly included in the input reviews.

\subsection{Stylistic Differences between Source and Target}
We investigate the stylistic differences between the input reviews and output summaries of the \textsc{ProSum} and \textsc{FewSum} datasets. Specifically, we measure the degree of relatedness between each word $w$ and a style $z$ by utilizing point-wise mutual information (PMI)~\cite{pavlick-nenkova-2015-inducing,kajiwara-2019-negative}, which is defined as follows:
\begin{align*}
    \text{PMI}(x, z) = \log\frac{p(w \mid z)}{p(w)}
\end{align*}
To handle potential sparsity issues, we applied Laplace smoothing to calculate PMI.

We present the top 20 words with the highest PMI values for each style $z$ in Tables~\ref{tab:pmi_prosum}-\ref{tab:pmi_fewsum_amazon}. As shown in Table~\ref{tab:pmi_prosum}, the Yelp style includes high PMI values for first-person pronouns such as "i", "my", and "me", whereas the Michelin point-of-view includes many expressions that are not commonly used in customer reviews such as "starring", "studded", and "brimming", indicating that training solely on Yelp reviews would not be sufficient for capturing the Michelin style.

Furthermore, we conduct a similar analysis on the \textsc{FewSum} datasets, as shown in Tables~\ref{tab:pmi_fewsum_yelp} and \ref{tab:pmi_fewsum_amazon}, and found that the human-written summaries on the \textsc{FewSum} also include many expressions that are not present in the input reviews. This indicates that there are stylistic differences between the input and output, even in \textsc{FewSum} datasets.
\begin{table}[ht]
    \centering
    \small
    \begin{tabular}{cc}
    \toprule
    Yelp & Michelin \\\midrule
    i & starring \\
    my & studded\\
    'm & brimming\\
    was & flaunts\\
    were & tailed\\
    me & tuck\\
    we & donning\\
    went & draws\\ 
    had & enriched\\
    came & talents\\
    amazing & bobbing\\
    our & black-and-white\\
    5 & towering\\
    ambiance & peruse\\
    4 & minimally\\
    am & thrill\\
    waiter & pressed-tin\\
    tried & tucking \\
    felt & golden-brown\\
    \bottomrule
    \end{tabular}
    \caption{Top 20 words with the highest PMI values for each style in the \textsc{ProSum} dataset.}
    \label{tab:pmi_prosum}
\end{table}

\begin{table}[ht]
    \centering
    \small
    \begin{tabular}{cc}
    \toprule
    Review & Summary \\\midrule
    i & generally \\
    my & particular\\
    we & well-liked\\
    our & features\\
    me & remarkably\\
    did & upscale\\
    'm & eyebrow\\
    've & impressive\\ 
    got & leaves\\
    he & general\\
    came & specializes\\
    us & regarded\\
    went & ratio\\
    again & payment\\
    'll & feature\\
    say & desired\\
    am & competent\\
    2 & well-regarded \\
    wo & tuesdays\\
    \bottomrule
    \end{tabular}
    \caption{Top 20 words with the highest PMI values for each style in the \textsc{Yelp} dataset of \textsc{FewSum}.}
    \label{tab:pmi_fewsum_yelp}
\end{table}

\begin{table}[ht]
    \centering
    \small
    \begin{tabular}{cc}
    \toprule
    Review & Summary \\\midrule
    my & dvds \\
    i & recommended\\
    had & versatile\\
    me & allows\\
    got & overall\\
    you & drawback\\
    am & tends\\
    've & ensure\\ 
    were & laptops\\
    our & weak\\
    'm & delicious\\
    he & generally\\
    her & child\\
    she & consumers\\
    said & drowsiness\\
    4 & adjusted\\
    week & fitting\\
    going & consistently \\
    see & offer\\
    \bottomrule
    \end{tabular}
    \caption{Top 20 words with the highest PMI values for each style in the \textsc{Amazon} dataset of \textsc{FewSum}.}
    \label{tab:pmi_fewsum_amazon}
\end{table}

\subsection{Pre-Training with Self-supervision}
\label{sub:pretrain_other}
We show the same analysis on Yelp and Amazon datasets. We observed the same trends with the \textsc{ProSum} dataset, showing the importance of pre-training with self-supervision across all three datasets used in the paper.
\begin{figure}[t]
    \centering
    \includegraphics[width=\linewidth]{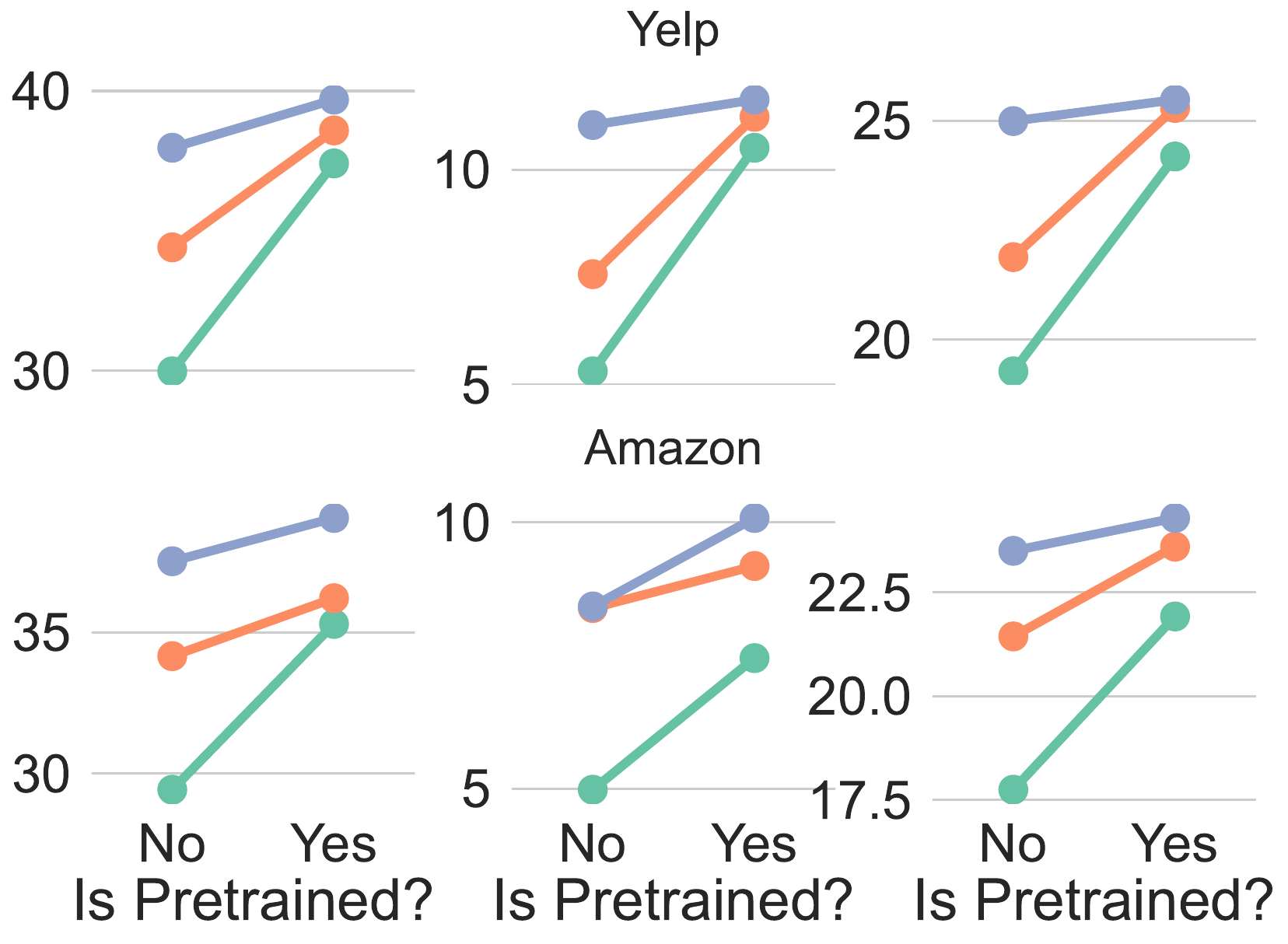}
    \caption{Comparison of summarization quality with and without pre-training on Yelp and Amazon datasets. The \textbf{\textcolor{bblue}{blue line}} denotes the model trained in a supervised setting, \textbf{\textcolor{oorange}{orange line}} denotes the model trained \textit{with} partial supervision and \textbf{\textcolor{ggreen}{green line}} denotes the model trained \textit{without} partial supervision. While pre-training with pseudo-training data improved the performance in all settings, we found a significant improvement, especially in the non-parallel settings (\textbf{\textcolor{oorange}{orange line}} and \textbf{\textcolor{ggreen}{green line}}).}
    \label{fig:ablation_other}
\end{figure}

\section{Qualitative Examples}
\label{sub:qualitative}
We present summaries of the \textsc{ProSum} data generated by Self-supervision (SS), Pipeline, SS + Noisy Pairing, and SS + Noisy Pairing + Partial Supervision in Table~\ref{tab:qualitative}.

For the self-supervised system (SS), the generated summary is a factually consistent summary with the source reviews, but it is a more review-like summary that includes first-person pronouns (e.g., I, my) and subjective opinions (e.g., \textit{The salmon skin hand roll and spicy tuna hand roll are two of my favorite things}).

Using the style transfer model (Pipeline), the generated summary contains attractive adjectives such as terrific, but the content of the summary cannot be changed by the style transfer model, so the summary still contains subjective opinions and first-person pronouns generated by the self-supervised system and introduce non-factual contents as well, e.g., \hl{oorange}{crispy pork was terrific}.

The model trained with the noisy paired dataset generates a more Michelin-like summary because it is fine-tuned with the same style of summaries.
However, because the noisy training pairs are used without partial supervision, the model generates a lot of non-factual content, such as the location of the restaurant (i.e., \hl{oorange}{San Francisco}) or the name of the chef (i.e., \hl{oorange}{Yoshihiko Kousaka}).

Finally, partial supervision (SS + Noisy Pairing + Partial Supervision) enabled the model to generate Michelin-like summaries while maintaining factual correctness, such as chef's name, \hl{ggreen}{Kiminobu Saito}.

\begin{table*}[t]
    \centering
    \footnotesize
    \begin{tabular}{p{\linewidth}}
        \toprule
        \napafont{Napa}\wine: Chef \hl{ggreen}{Kiminobu Saito} has been at this spot for years , and it 's no surprise that he 's become one of the city 's most beloved chefs . His attention to detail is evident in every single detail of this restaurant , from \hl{ggreen}{the cleanliness of the dining room} to \hl{ggreen}{the attentive service} . \hl{ggreen}{The omakase is the only way to go here} , and the quality of the fish is just as important as the service . \hl{ggreen}{Start your meal with a cup of miso soup} , \hl{ggreen}{then move on to a selection of handrolls} , like \hl{ggreen}{the salmon skin hand roll and the toro , which are both tender and flavorful }. The chef 's personal touch is also evident in the presentation of each piece of fish , and you 'll be hard pressed to find a better one than the one on offer here .\\\midrule
        \textbf{w/o Partial Supervision: } If you 're looking for a laid-back , family-friendly sushi bar in the heart of downtown \hl{oorange}{San Francisco} , this is the place to be . \hl{ggreen}{The dining room may be small} , but \hl{ggreen}{the mood is upbeat} and \hl{ggreen}{the service is warm and friendly} . Chef \hl{oorange}{Yoshihiko Kousaka} is the man behind the counter , and his skillfully curated omakase is evident in every morsel of food he prepares . The quality of each morsel is impeccable , and you can tell that he takes great pride in every detail of his work-whether it 's \hl{ggreen}{hand-rolls} , \hl{oorange}{hand-formed karaage} , or \hl{oorange}{hand-tossed morsels of fish in a wood-fired oven} . Highlights include a trio of handrolls made with house-made sashimi , each with its own unique flavor profile and texture , and a hand-ground toro that pairs perfectly with a glass of sake or sake-battered sake .\\\midrule
        \textbf{w/o Noisy Pairing and Partial Supervision: }Sushi Note is one of the best sushi restaurants I 've ever been to in my life . The service is impeccable , the food is amazing , and Chef \hl{ggreen}{Kiminobu Saito} is an absolute pleasure to talk to . The omakase is the way to go if you want to experience \hl{ggreen}{the full experience of sushi and wine pairings . The salmon skin hand roll and spicy tuna hand roll are two of my favorite} \hl{ggreen}{things on the menu .} \hl{oorange}{I 'm not a huge fan of spicy tuna , but the spicy tuna was so good that I had to order it again the } \hl{oorange}{next time I went} . \hl{ggreen}{The scallop with truffle truffle and toro is also a must try} . If you 're in the mood for sushi , this is the place to go . It 's a little pricey , but it 's worth it for the quality of the food and the service .\\\midrule
        \textbf{Pipeline: } In fact , I 've never been to a better sushi bar in my life . The service is terrific , the food is terrific , and Chef \hl{ggreen}{Kiminobu Saito} is a terrific talker . Once inside , \hl{oorange}{order a cocktail} and \hl{ggreen}{admire the full sushi and wine experience} . \hl{ggreen}{The salmon roll and spicy tuna hand roll are my favorite} . Do n't like spicy tuna , but the \hl{oorange}{crispy pork was terrific} . \hl{oorange}{Starters like truffle and truffle are also a must try with these truffle and truffle} . It 's the right place to go to the sushi counter . It 's worth every second for this quality of the food and the service .\\
        \bottomrule
    \end{tabular}
    \caption{Qualitative examples on \textsc{ProSum} dataset. Faithful/unfaithful contents are highlighted in \hl{ggreen}{green}/\hl{oorange}{orange}.}
    \label{tab:qualitative}
\end{table*}

\section{More details of human evaluation}
\label{sec:human_eval_appendix}
We performed human evaluation using the Appen platform.\footnote{\url{https://appen.com/}} We sampled 50 instances from the \textsc{ProSum} test set and recruited three crowd workers for each instance to evaluate the summaries generated by four systems: Self-supervision, Pipeline, \napafont{Napa}\wine without Partial Supervision, and \napafont{Napa}\wine. The summaries and their corresponding reviews were presented to the worker in a random order, and the workers judged them using a 4-point Likert scale. The workers were asked to judge the summaries based on the following criteria:

\textit{Fluency}: the summary should be grammatically correct and easy to read; \textit{Relevance}: the summary should be consistent with the input reviews; \textit{Attractiveness}: the likelihood of the summary being shown on a professional restaurant website, such as Michelin Restaurant Guide.

\begin{figure*}[t]
    \centering
    \includegraphics[width=\linewidth]{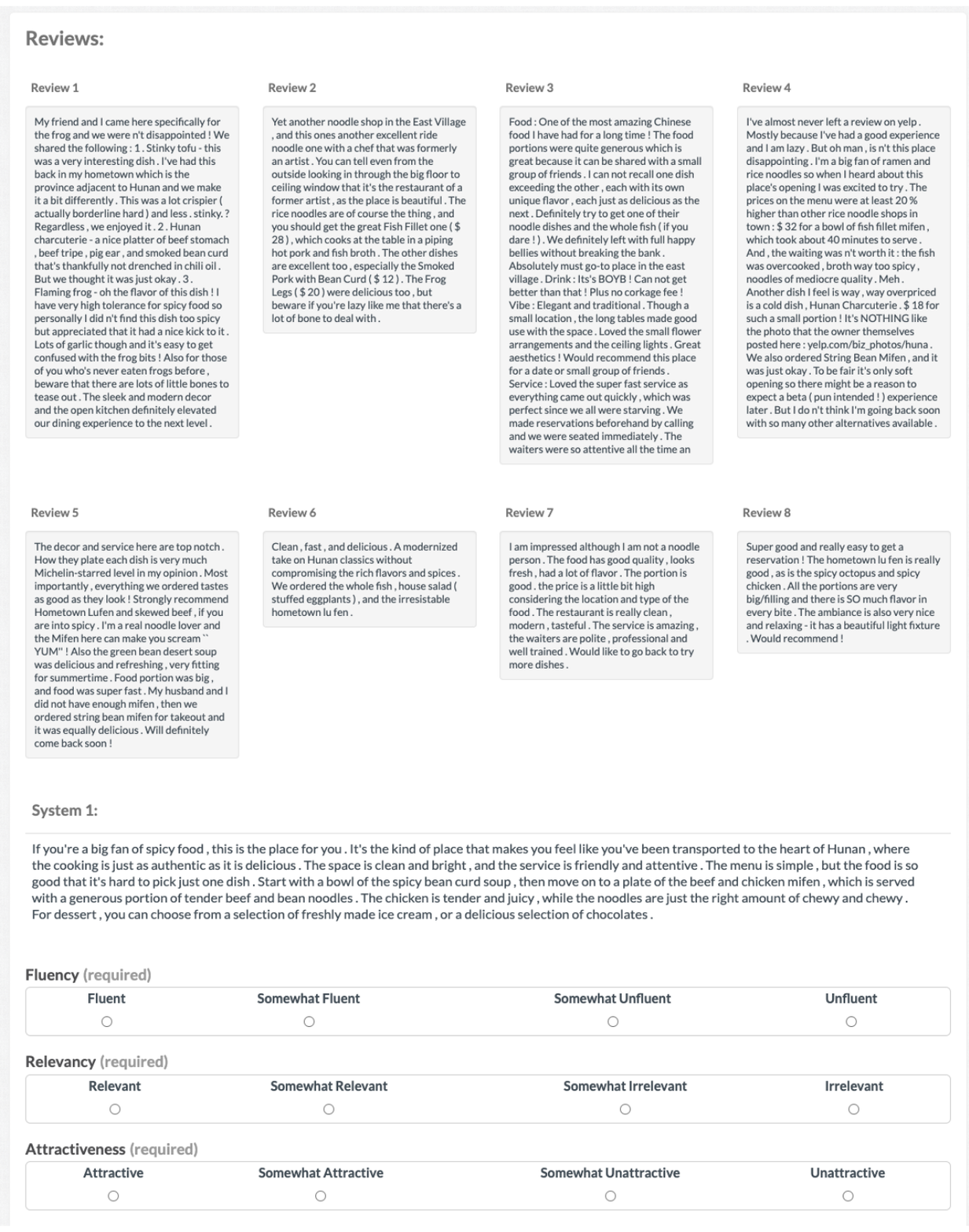}
    \caption{Human evaluation task}
    \label{fig:screen}
\end{figure*}

We also show the annotation screen in Figure~\ref{fig:screen}. The annotators are asked to select three aspects of summaries based on the system's generation.
The inter-annotator agreement was measured using Krippendoff's alpha~\cite{Krippendorff1980KrippendorffKC}, which was 0.456 for fluency, 0.458 for relevance, and 0.338 for attractiveness.

\bibliography{custom}